\pdfoutput=1

\documentclass[11pt]{article}

\usepackage[final]{acl}

\usepackage{times}
\usepackage{latexsym}

\usepackage[T1]{fontenc}

\usepackage[utf8]{inputenc}

\usepackage{microtype}

\usepackage{inconsolata}

\usepackage{graphicx}

\usepackage{colortbl}
\PassOptionsToPackage{table}{xcolor}

\usepackage{booktabs}       
\usepackage{amsfonts}       
\usepackage{nicefrac}       
\usepackage{microtype}      
\usepackage{xcolor}         

\usepackage{graphicx}
\usepackage{adjustbox}
\usepackage{makecell}
\usepackage{caption}
\usepackage{multirow}
\usepackage{subcaption}
\usepackage{comment}
\usepackage{svg}
\usepackage{amsmath}
\usepackage{enumitem}

\def\checkgray{\textcolor{gray}}

\title{\textsc{ComplexTempQA}: A 100m Dataset for Complex Temporal Question Answering}

\author{
    \textbf{Raphael Gruber\textsuperscript{1}, Abdelrahman Abdallah\textsuperscript{1}, Michael F{\"a}rber\textsuperscript{2}, Adam Jatowt\textsuperscript{1}} \\
    \textsuperscript{1}University of Innsbruck, \textsuperscript{2}Technical University Dresden \\
    \texttt{\{ra.gruber, abdelrahman.abdallah, adam.jatowt\}@uibk.ac.at,} \\
    \texttt{michael.faerber@tu-dresden.de}
}

\begin{document}
\maketitle
\begin{abstract}
We introduce \textsc{ComplexTempQA},\footnote{Dataset and code available at: \url{https://github.com/DataScienceUIBK/ComplexTempQA}
} a large-scale dataset consisting of over 100 million question-answer pairs designed to tackle the challenges in temporal question answering. 
\textsc{ComplexTempQA} significantly surpasses existing benchmarks  in scale and scope. Utilizing Wikipedia and Wikidata, the dataset covers questions spanning over two decades and offers an unmatched scale. We introduce a new taxonomy that categorizes questions as \textit{attributes}, \textit{comparisons}, and \textit{counting} questions, revolving around events, entities, and time periods, respectively. A standout feature of \textsc{ComplexTempQA} is the high complexity of its questions, which demand reasoning capabilities for answering such as across-time comparison, temporal aggregation, and multi-hop reasoning involving temporal event ordering and entity recognition. Additionally, each question is accompanied by detailed metadata, including specific time scopes, allowing for comprehensive evaluation of temporal reasoning abilities of large language models. 
\end{abstract}



\maketitle

\section{Introduction}
\begin{figure}[t]
    \centering
\includegraphics[width=.5\textwidth]{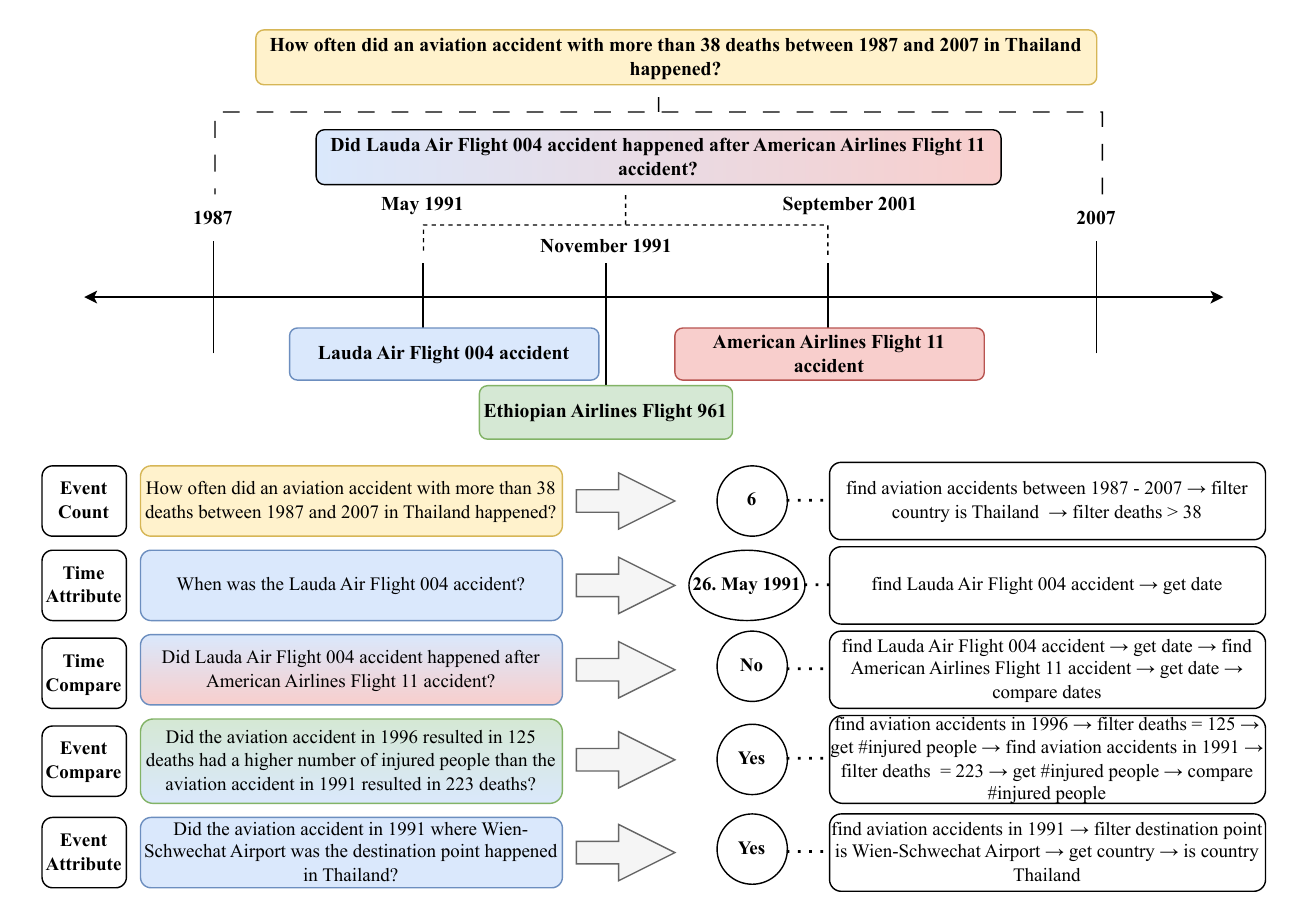}
   \caption{Example types of temporal reasoning in questions sampled from \textsc{ComplexTempQA} (left) with the required inference steps (right) and timeline-based visualization (above).}
    \label{fig:timeline}
\end{figure}

Temporal Question Answering (TQA) refers to answering questions that require both understanding of and reasoning about temporal knowledge~\cite{DBLP:conf/www/Jatowt22,jia2018tempquestions,TEQUILA,TORQUE}. This sets TQA apart from traditional Question Answering (QA). 
%
%
Developing effective TQA solutions naturally requires effective and challenging datasets.
Existing TQA datasets, such as TORQUE~\cite{TORQUE}, TEQUILA~\cite{TEQUILA}, ArchivalQA~\cite{ArchivalQA}, and ChroniclingAmericaQA~\cite{DBLP:conf/sigir/PiryaniMJ24}, fall however short in several respects:
\textit{First}, they are limited in size, typically containing only a few thousand questions. This poses significant challenges for effectively training LLMs, as illustrated in~\cite{TORQUE,TEMPQuestions,QATempEval,OngSSG23,WeiSMYLZZL23,NaikBR19}.
\textit{Second}, while these datasets predominantly focus on specific question categories related to entities or time periods, they lack comprehensive coverage of a wide variety of question types.
\textit{Third}, they generally include only straightforward questions and omit complex inquiries that require multi-step inference to generate accurate responses, thereby limiting the ability of trained models to handle temporal reasoning tasks~\cite{TORQUE,OngSSG23,WeiSMYLZZL23,NaikBR19}.
\textit{Fourth}, the prior datasets 
lack features such as popularity scores, which could indicate the relative ease of answering questions. They also do not allow filtering by specific time periods, a capability essential for detailed and customized temporal studies.

To address these challenges, we introduce \textsc{ComplexTempQA}, a novel dataset that surpasses existing resources in both scale and complexity. It is particularly suitable for the analysis, training, and evaluation of LLMs and QA systems on complex temporal knowledge. Our dataset offers four key contributions:

\begin{enumerate}[leftmargin=.5cm]
 \item \textit{Scale:} \textsc{ComplexTempQA} comprises over \textit{100 million question-answer pairs}, making it by far the largest dataset available for Temporal QA.

 \item \textit{Question Types and Taxonomy:} The dataset includes diverse categories of questions, such as \textit{attribute}-, \textit{comparison}-, and \textit{counting}-type questions, each pertaining to events, entities, or time periods. Questions are generated at scale based on facts from Wikidata and representative question patterns identified from Wikipedia, ensuring broad coverage of domains and question types. Moreover, we employ the IPTC Media Topics taxonomy to diversify thematic scope.

 \item \textit{Complexity and Temporal Scope:} Questions in \textsc{ComplexTempQA} require advanced reasoning skills, including event/entity matching, multi-hop inference, cross-time comparisons, and temporal ordering. 

\item \textit{Metadata and Evaluation:} Each question is enriched with detailed metadata, including the relevant time period within the overall dataset's time span from 1987 to 2023 and a popularity score that categorizes questions as popular or unpopular based on the type of question and the anticipated familiarity with the entities involved. This metadata allows for precise training and evaluation of language models, specifically concerning their ability to adapt to varying temporal contexts over time.
\end{enumerate}

\textsc{ComplexTempQA} serves multiple purposes in advancing the study and application of LLMs in reasoning over temporal factual knowledge. Primarily, it enables a thorough analysis of LLM performance by addressing the need to understand temporal factual knowledge, identify temporal blind spots, and assess temporal reasoning capabilities~\cite{wallat,DBLP:journals/corr/abs-2308-00002}. Beyond performance evaluation, the dataset provides a foundational platform for the development of advanced question-generation tools. Our structured taxonomy facilitates the refinement and creation of specialized taxonomies for targeted tasks and datasets. Moreover, specific subsets of \textsc{ComplexTempQA} can be filtered to focus on particular types of temporal questions or particular desired time frames, supporting targeted research.

A key component of our work is evaluating how state-of-the-art LLMs handle the complex temporal questions posed by \textsc{ComplexTempQA}. In Section 6, we benchmark a range of models using various approaches, including zero-shot, few-shot, and retrieval-augmented generation (RAG). These evaluations offer insights into the current capabilities and limitations of LLMs in processing temporal information.

Overall, we make the following contributions:

\noindent
\begin{enumerate}[leftmargin=*]
    \item We introduce and publicly release \textsc{ComplexTempQA}\footnotemark[1], a large-scale dataset comprising over 100 million question-answer pairs for temporal question answering, structured around a novel taxonomy of temporal question types.
    
    \item We detail our methodology for dataset creation, which includes data retrieval from Wikidata, event extraction from Wikipedia, rigorous complexity assessment and rating, as well as filtering based on temporal ambiguity. Moreover, the dataset can be easily extended to cover additional time periods.
    
    \item We benchmark diverse LLMs on \textsc{ComplexTempQA} using zero-shot, few-shot prompting, and retrieval-augmented generation (RAG) approaches, providing  insights into their performance on temporal question answering tasks of varying complexity.
\end{enumerate}


\section{Related Work}


\begin{table*}[tb]
\centering

\large
\renewcommand{\arraystretch}{1.3} 
\begin{adjustbox}{width=\textwidth,center}
\begin{tabular}{
    l c r
    >{\centering\arraybackslash}m{2.5cm}
    >{\centering\arraybackslash}m{2.2cm}
    >{\centering\arraybackslash}m{2.5cm}
    >{\centering\arraybackslash}m{3.5cm}
    >{\centering\arraybackslash}m{2.5cm}
    >{\centering\arraybackslash}m{2.5cm} c c
}
\toprule
\textbf{Dataset} & \textbf{\shortstack{Temporal\\Questions}} & \textbf{\#Questions} & \textbf{\shortstack{Creation\\Method}} & \textbf{Source} & \textbf{\shortstack{Answer\\Type}} & \textbf{\shortstack{Complex\\Question Types}} & \textbf{\shortstack{Question\\Type}} & \textbf{Time Frame} & \textbf{\shortstack{Temporal\\Metadata}} & \textbf{Multi-Hop} \\ 
\midrule
NewsQuizQA \cite{abs-2102-09094} & No & 20K & \shortstack{Crowd\\sourced} & News & \shortstack{Multiple\\choice} & - & Attr & 2018--2022 & No & No \\[1ex]
\rowcolor{gray!10} NewsQA \cite{TrischlerWYHSBS16} & Partially & 119K & \shortstack{Crowd\\sourced} & News & Extractive & TimeAttr & Attr & 2007--2015 & No & No \\[1ex]
HOTPOTQA \cite{HotpotQA} & No & 113K & \shortstack{Crowd\\sourced} & Wikipedia & Extractive & - & Attr, Comp & Unspecified & No & 2 hops \\[1ex]
\rowcolor{gray!10} LC-QuAD 2.0 \cite{LC-QuAD2} & Partially & 30K & \shortstack{Crowd\\sourced} & Wikipedia & Extractive & TimeAttr & Attr, Count & Unspecified & No & No \\[1ex]
TORQUE \cite{TORQUE} & Yes & 21K & \shortstack{Crowd\\sourced} & News & Generative & TimeAttr & Attr & Unspecified (short) & No & No \\[1ex]
\rowcolor{gray!10} Time-Sensitive-QA \cite{Time-SensitiveChen} & Yes & 41K & \shortstack{Aut.\\Generated} & Wikipedia & Extractive & TimeAttr & Attr & Unspecified (long) & No & No \\[1ex]
TempQuestions \cite{TEMPQuestions} & Yes & 1K & \shortstack{Aut.\\Generated} & Freebase & Extractive & \shortstack{TimeAttr,\\TimeComp,\\TimeCount} & \shortstack{Attr, Comp,\\Count} & History & No & No \\[1ex]
\rowcolor{gray!10} TKGQA \cite{OngSSG23} & Yes & 5K & \shortstack{Aut.\\Generated} & News & Extractive & \shortstack{TimeAttr} & Attr & 2022 & No & No \\[1ex]
MenatQA \cite{WeiSMYLZZL23} & Yes & 2K & \shortstack{Aut.\\Generated} & Wikipedia & Extractive & TimeAttr & Attr & Unspecified (long) & No & No \\[1ex]
\rowcolor{gray!10} TDDiscourse \cite{NaikBR19} & Yes & 6K & \shortstack{Aut.\\Generated} & News & Extractive & TimeAttr & Attr & Unspecified (short) & No & No \\[1ex]
ArchivalQA \cite{ArchivalQA} & Partially & 532K & \shortstack{Aut.\\Generated} & News & Extractive & TimeAttr & Attr, Count & 1987--2007 & No & No \\[1ex]
\rowcolor{gray!10} \textbf{\textsc{ComplexTempQA}} & 
\textbf{Yes} & 
\textbf{100,228K} & 
\shortstack{\textbf{Aut.}\\\textbf{Generated}} & 
\textbf{Wikipedia} & 
\shortstack{\textbf{Extractive,}\\\textbf{Boolean}} & 
\shortstack{\textbf{TimeAttr,}\\\textbf{TimeComp,}\\ \textbf{TimeCount,}\\\textbf{Unnamed questions}} & 
\shortstack{\textbf{Attr,}\\\textbf{Comp,}\\\textbf{Count}} & \textbf{1987--2023} & 
\textbf{Yes} &
\textbf{$\leq$ 2 hops} \\
\bottomrule
\end{tabular}
\end{adjustbox}
\caption{Comparison of \textsc{ComplexTempQA} with existing datasets highlighting the key aspects of question creation methodologies, answer types, complexity, temporal scope, and structure. Attr means \textit{attribute}-type question, Comp denotes \textit{comparison}-type questions and Count are \textit{counting}-type questions.}
\label{tab:related-work}
\end{table*}

Table \ref{tab:related-work} gives an overview of question answering datasets 
showing a notable discrepancy in question volume, with our dataset substantially surpassing others. 
While several TQA datasets contain complex temporal questions, they largely revolve around \textit{TimeAttr} (time attribute) inquiries, which focus on relatively simple questions such as ones about the duration and order of events or entities. 
TORQUE~\cite{TORQUE} introduces questions that emphasize temporal relations such as ``before,'' ``after,'' and ``start.'' The authors trained a model to evaluate questions specifically based on these temporal constraints. QA TempEval dataset~\cite{QATempEval} has been designed with a focus on temporal entities and relations, which are easier to generate automatically. TEQUILA~\cite{TEQUILA} uses temporal expressions like dates or implicit temporal signals such as ``before'' or ``after.''

In addition to these works, TempQuestions~\cite{jia2018tempquestions} has been released as a benchmark for temporal questions, containing 1,271 questions that are all temporal in nature, paired with their answers. That work provides a simple definition for temporal questions and demonstrates the need for further research on complex queries. 
Stricker~\cite{stricker2023question} applies answer extraction techniques from general question answering to retrieve temporal answers by identifying and processing time expressions. Their approach focuses on structured temporal information and distinguishes between absolute and relative time expressions. 

Unlike other datasets, ours stands out due to its unique characteristics: \textbf{(a)} it comprises a number of question-answer pairs that is orders of magnitude larger than those in other datasets, \textbf{(b)} it categorizes these questions into specific types, \textbf{(c)} it includes complex questions, and \textbf{(d)} the questions are temporal in nature, with each strictly assigned to a specific time span.




\section{Dataset Characteristics}

We describe the \textsc{ComplexTempQA} characteristics along the four dimensions.

\textbf{Size:~}
\textsc{ComplexTempQA} comprises 100 million question-answer pairs and covers the period from 1987 to 2023. 
The dataset has been curated to probe the understanding of temporal knowledge within this 36-year span, encapsulating events, entity milestones, and other time-sensitive data.

\begin{figure}[tbp]
    \centering
    \begin{minipage}{0.20\textwidth}
        \centering
        \scriptsize
  
        \begin{adjustbox}{width=\textwidth}
        \begin{tabular}{lr}
            \toprule
            \textbf{Name} & \textbf{Total} \\
            \midrule
            Attribute Event & 83,798 \\
            Attribute Entity & 84,079 \\
            Attribute Time & 9,454 \\
            Comparison Event & 25,353,340 \\
            Comparison Entity & 74,678,117 \\
            Comparison Time & 54,022,952 \\
            Counting Event & 18,325 \\
            Counting Entity & 10,798 \\
            Counting Time & 12,732 \\
            \midrule
            Multi-Hop: & 76,933 \\
            Unnamed Event: & 8,707,123 \\
            \textbf{Total:} & \textbf{100,228,457} \\
            \bottomrule
        \end{tabular}
        \end{adjustbox}
              \captionof{table}{Dataset distribution (time questions are integrated within events or entities).}
        \label{tab:Entities1b}
    \end{minipage}
    \hfill
    \begin{minipage}{0.25\textwidth}
        \centering

        \includegraphics[width=0.9\textwidth]{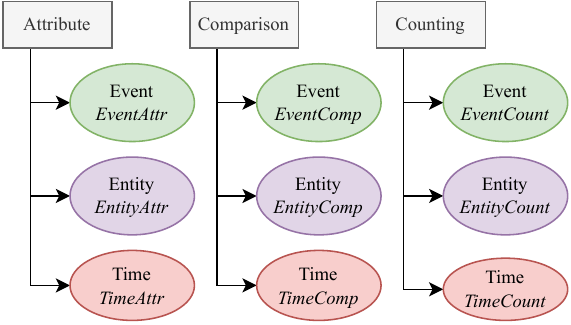}
        \caption{Taxonomy of temporal question types.}
        \label{fig:taxonomy}
    \end{minipage}
\end{figure}

\textbf{Question Types and Taxonomy:} 
The primary objective when constructing \textsc{ComplexTempQA} was to incorporate a broad spectrum of temporal questions. 
Our dataset offers extensive coverage across diverse subjects by aligning with IPTC MediaTopic standards \footnote{\url {https://iptc.org/standards/media-topics/}}, ensuring a broad range of relevant topics. 
The questions are also organized using a taxonomy that categorizes them by the questioned entity or event and the nature of the knowledge they probe, divided into \textit{attribute}, \textit{comparison}, and \textit{counting} questions (see Figure~\ref{fig:taxonomy}). The numbers for each question type are shown in Table \ref{tab:Entities1b}. Below we provide description of each type, while specific examples are given in 
Appendix A.

\textit{Attribute}-type questions ask about the properties of events or entities, or relate to a specific time (e.g., \textit{``When was the fall of the Berlin Wall?''}). Answers usually consist of \textbf{events}, \textbf{entities}, \textbf{dates} (or lists of dates), or \textbf{real numbers}. For numerical answers involving units, we convert all measurements to the International System of Units (SI units) to ensure consistency. 

\textit{Comparison}-type questions involve the comparison of up to three events, entities, or time periods. The comparison aspect can be either numerical or temporal. For example, we compare two temporal attributes in the question: \textit{``Did Halabja chemical attack happen after John F. Kennedy Jr. plane crash?''}. The answer to these questions can be either \textbf{true}, \textbf{false}, an \textbf{entity}, or an \textbf{event}. 
Due to the large number of possible questions that can be created based on comparison, this type is the most frequent in our dataset. We note that comparative questions require more reasoning than relatively simpler attribute type questions.


\textit{Counting}-type questions—asking about a form of aggregation—record the frequency of a particular event type or the occurrence of an attribute for an entity. An example is \textit{``How often did an aviation accident with more than 38 deaths occur in Thailand between 1987 and 2023?''} To make the answering of \textit{counting}-type questions more precise, we always consider a specific time period and, in some cases, an attribute threshold. The threshold is determined by calculating the average value of the attribute across all comparable instances. For example, in the given example, the threshold refers to the number of deaths associated with the event. The answer to these questions is always a \textbf{natural number}.


\textbf{Popularity of Questions.~}
\label{sec:Rating}
Questions are categorized as either \textbf{popular} or \textbf{unpopular} based on a popularity score derived from the English Wikipedia's page view statistics and the intrinsic complexity of the questions (see Sec.~\ref{sec:dataset-source-extraction}). A question is considered \textit{popular} if all of its constituent entities or events are rated as common—that is, they have high page view counts that reflect broad public familiarity. 
We use the standard deviation for thresholds since page views follow a long-tail distribution. An average-based threshold would misclassify low-view questions as well-known. This approach helps ensure that notable events and entities are classified as popular.
For example, a popular \textit{time attribute}-type question is: \textit{"When was the death of Diana, Princess of Wales?"} 

Questions that require advanced reasoning—such as \textit{counting}-type questions, multi-hop queries, or modified versions of standard questions that implicitly reference an event without explicitly naming it (which we refer to as \textbf{unnamed event questions}; see Sec.~\ref{sec:complexity} for details)—are inherently more challenging and are automatically classified as \textit{unpopular}. 

\textbf{Metadata and Evaluation.~}
\textsc{ComplexTempQA} includes several metadata fields for each question, serving multiple purposes such as facilitating the retrieval of entities and events from the question or answer, enabling in-depth analysis of the dataset, and supporting segmentation based on attributes such as type, year, or popularity score.
Specifically, the metadata comprises the following:
\begin{itemize}[leftmargin=*]
    \item \textbf{Type of Question:} The question type is specified based on the taxonomy shown in Figure~\ref{fig:taxonomy}.
    \item \textbf{Identifiers:} All corresponding Wikidata item identifiers for the entities or events that are either the subject of the question or part of the answer are included.
    \item \textbf{Country:} The country associated with the questioned entities or events and their answers is provided, if applicable.
    \item \textbf{Complexity Indicators:} These indicate properties used for generating multi-hop questions and whether the question is an unnamed event question.
    \item \textbf{Popularity Rating:} The popularity of the question is rated based on whether it concerns popular or unpopular entities or events.
    \item \textbf{Time Span:} A temporal range is specified for each question.
\end{itemize}
Appendix B provides further details on each metadata field together with an illustrative example.

\section{Dataset Creation Pipeline}

\begin{figure}[t]
    \centering
\includegraphics[width=.46\textwidth]{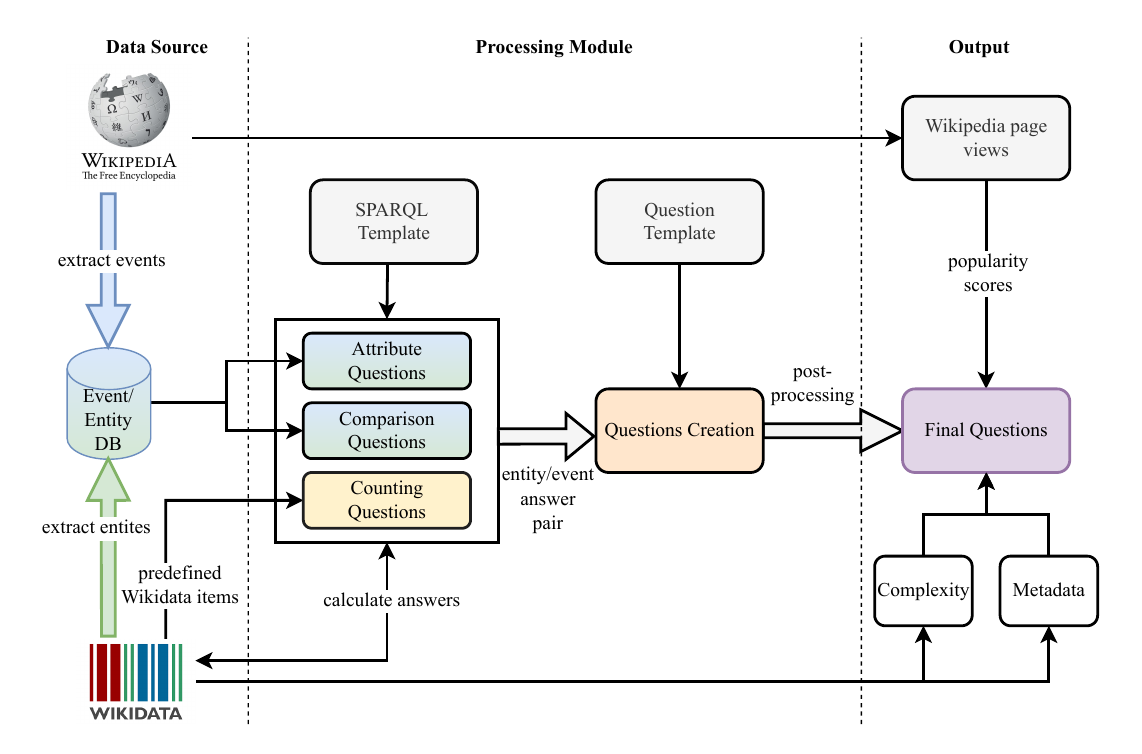}
   \caption{A simplified view of the dataset creation pipeline, showing how events and entities from Wikipedia and Wikidata are processed into final question–answer pairs with metadata and popularity scores.}
    \label{fig:pipeline}
\end{figure}

In this section, we present our methodology for constructing the dataset. We first describe the data acquisition process from Wikipedia and Wikidata, followed by the procedure for generating multi-hop questions. Finally, we detail the formation of question-answer pairs. An overview of the entire pipeline is illustrated in Figure \ref{fig:pipeline}.

\subsection{Dataset Source Extraction}
\label{sec:dataset-source-extraction}

Different question types call for distinct data sources to best capture the required information while ensuring reliability. 
For the \textit{event attribute}-type questions, we extracted every entry from ``single year'' Wikipedia pages (see, for example, year 1987\footnote{\url{https://en.wikipedia.org/wiki/1987}}), collecting information on significant events occurring over a 36-year span (1987--2023). 
We then applied a filtering step to discard entries lacking a clear timestamp or falling outside our time frame. 

We used the same Wikipedia sources to generate \textit{comparison}-type questions, comparing up to three events (or entities) by examining attributes such as date, location, or other numerical values. These questions typically include terms like ``higher/lower,'' ``before/after,'' or ``happened first.''

To construct \textit{entity attribute}-type questions, we compiled a list of entities by querying Wikidata through SPARQL, focusing on items such as ``movies'' and ``heads of state.'' These predefined items were derived from the IPTC MediaTopic standards
and manually curated. Because each item can yield an extensive list of possible entities, we filtered results based on the Wikipedia page views of those entities. For example, a raw list of over 100,000 ``movie'' entities was reduced to 35,991 by applying a page-view threshold.

Lastly, for \textit{counting}-type questions, we used a distinct approach, as these questions enumerate sets of events or entities instead of focusing on a single event or entity. We began by creating SPARQL query templates tailored to specific counting tasks. For instance, by querying the Wikidata item ``Presidents of the United States'' within a specific time range, we obtained a list of all relevant individuals for potential \textit{counting}-type questions. In some cases, we could directly use the resulting list; however, certain lists might be incomplete if Wikidata omits minor events (e.g. earthquakes). To mitigate such omissions, we introduced a threshold. For example, for ``number of deaths in some calamity,'' we used the average values of corresponding attributes from Wikidata to define the cutoff.

\subsection{Dataset Complexity Enhancement}
\label{sec:complexity}

We next increased the complexity of the dataset by introducing a module to create \textit{multi-hop questions}. The idea is to leverage shared properties across the events and entities extracted in the previous step, requiring a multi-step reasoning processes to answer a question.

\paragraph{Example of Multi-Hop Question.}

Consider the question:
\begin{quote}
\textit{``What was the highest point of the country where the 1988 Summer Olympics happened, in meters?''}
\end{quote}

\begin{enumerate}[noitemsep]
    \item Select a specific event (the \emph{1988 Summer Olympics}) as the initial anchor.
    \item Retrieve a property from that event (\emph{country}).
    \item Formulate a further query about that country by selecting an additional attribute (the \emph{highest point}, in this case).
\end{enumerate}

This multi-hop process requires multiple pieces of information across different domains---first about an event (location, date), then about a geographical feature related to that location. Multiple hops necessitate more complex computation which can be especially challenging in temporal settings when asking about more obscure events or entities from the past.

As another complexity enhancement, we created additional event references with implicit naming to expand the range of questions. For instance, ``Lauda Air Flight 004 accident'' was rephrased as ``the aviation accident in 1991 which resulted in 223 deaths.'' We formed such references by combining the event’s year with a property (e.g., number of fatalities), then confirming via a SPARQL query that no other events shared these attributes to ensure the lack of temporal ambiguity. 

\subsection{Dataset Formation and Enrichment}

The final step involved constructing the actual question--answer pairs by integrating events, entities, and counting results into predefined question templates (see Appendix~\ref{sec:appendixTemplates}). 
Designing the templates for generating question-answer pairs presented several challenges. The templates needed to be general enough to cover a broad range of temporal reasoning tasks while remaining structured enough to maintain coherence and logical validity. The dataset had to support diverse temporal expressions, such as absolute dates, relative references, and temporal signals, requiring precise formatting and adaptability. Ensuring the generated questions were grammatically correct and naturally phrased was crucial, necessitating careful design to avoid awkward sentence constructions. Some questions required multiple reasoning steps involving different events and entities, making it challenging to construct templates that preserved logical consistency while maintaining clarity. 

Below is an overview:

\begin{itemize}[leftmargin=*]
    \item \textit{Attribute-type questions}: Query specific attributes of an event or entity. Following~\cite{Time-SensitiveChen}, we also included additional relations to enhance the \emph{comprehensiveness} of these queries.
    \item \textit{Comparison-type questions}: Select up to three events or entities to compare, using terms such as ``higher/lower,'', ``smallest/largest,'' ``before/after,'' or ``happened first.'' 
    \item \textit{Counting-type questions}: Specify a time frame (e.g. five years) and query the relevant category (e.g. earthquakes, Nobel Peace Prize recipients) to count the number of matching events or entities.
\end{itemize}

Once the QA pairs were generated, we enriched them with \textit{metadata} such as corresponding Wikidata IDs, the relevant country, and the specific time frame of the question. We additionally assigned \textit{popularity scores} as described in Section~\ref{sec:Rating}. 

\begin{figure}[htbp]
    \centering
    \begin{minipage}[t]{0.5\textwidth}
         \centering
          \vspace{0pt}
          \includegraphics[width=\textwidth]{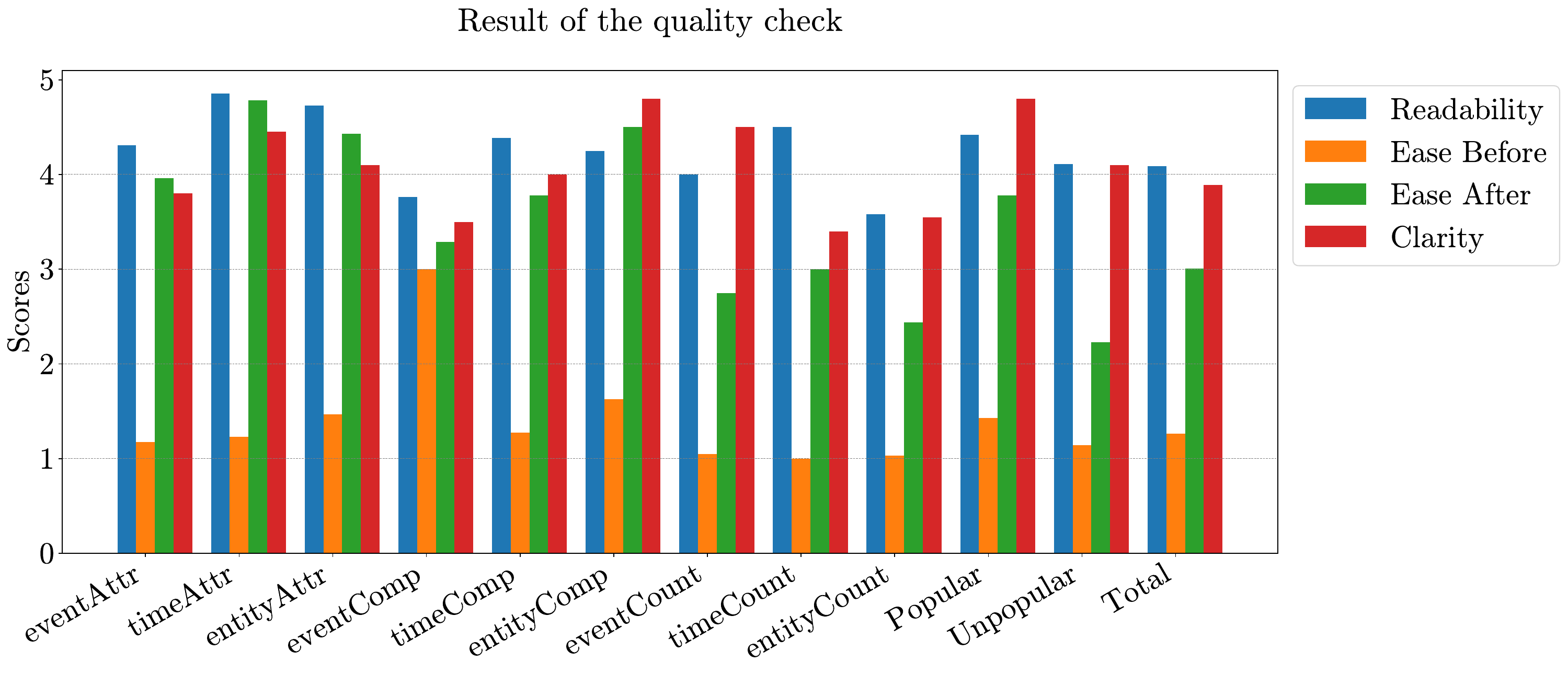}
          \vspace{-0.6cm}
          \caption{Result of the quality check.}
          \label{fig:qualitycheck}
    \end{minipage}
\quad 
    \begin{minipage}[t]{0.42\textwidth}
        \centering
        \captionsetup{type=table} 
          
\begin{adjustbox}{width=\textwidth,center}
\begin{tabular}{lcccccc}
\hline
Method & Parameters & Precision  & Recall  & F1  & Con \\ \hline
Zephyr & 7B & 3.76 & 33.50 & 4.90 & 55.97\\  
Falcon & 7B & 0.31 & 34.22 & 0.62 & \textbf{64.21} \\  
Llama-chat 7B & 7B & 3.68 & 33.94 & 6.09 & 50.32\\  
Mistral & 7B & 3.73 & \textbf{48.34} & 6.33 & 58.31 \\  
LLama-chat 13B& 13B & 3.61 & 32.77 & 6.05 & 53.27 \\  
Vicuna & 33B & 2.02 & 37.27 & 3.63 & 58.26 \\  
Mixtral & 8x7B & 3.34 & 40.41 & 5.65 & 60.21 \\  
LLama-chat 70B& 70B & \textbf{5.19} & 39.30 & 8.31 & 57.33 \\   
Wizardlm & 70b & 1.63 & 27.63 & 2.80 & 50.44 \\ 
GPT-3.5 & - & 2.68 & 29.12 & 4.56 & 46.79 \\ 
GPT-4-o & - & \textbf{8.30} & 29.90 & \textbf{10.47} & 46.55 \\ 
\hline
\end{tabular}
\end{adjustbox}
          \caption{Performance of zero-shot LLMs.}

\label{tab:zero-shotmetrics_llm}

    \end{minipage}
\end{figure}

\section{Dataset Quality Assessment}

The quality assessment by human raters is essential to ensure that \textsc{ComplexTempQA} meets the high standards of clarity and readability required for advanced research. Expert evaluation allows us to detect and address potential errors, ambiguities, or biases, thereby reinforcing the dataset’s reliability.

To evaluate \textsc{ComplexTempQA}, we conducted a user study with 11 volunteers (4 females, 7 males; 4 with secondary education and 7 with postgraduate degrees; ages 26–56, predominantly in their 20s and 30s). Participants assessed 450 randomly selected questions, evenly distributed across all types, using a 5-point Likert scale across four dimensions: (1) \textbf{readability}; (2) \textbf{ease of answering} \textit{prior to using a web search engine}; (3) \textbf{ease of answering} \textit{after conducting a web search}; and (4) overall \textbf{clarity} (including lack of ambiguity). For dimensions (2) and (3), participants first rated the questions based solely on their own knowledge (2), and then again after performing web search (3) with explicit instructions not to use any large language models. 

Figure~\ref{fig:qualitycheck} shows that questions generally received high ratings for readability and clarity. However, participants reported that answering questions without a search engine was challenging, while the access to web search results significantly improved answerability. Notably, \textit{counting}-type questions remained the most difficult to answer even after web searches, as they often require special type of inference. Additionally, some \textit{counting}- and \textit{comparison}-type questions received lower clarity scores, suggesting opportunities for further refinement. Overall, the expected familiarity of the questions significantly influenced the ratings, with questions classified as unpopular proving more challenging both before and after web search. 

\section{Experiments}
We evaluate multiple Large Language Models (LLMs), including Llama, Mistral, Mixtral, Falcon, Vicuna, Zephyr, WizardLM, GPT-3.5, and GPT-4-o, each selected for its unique strengths. We discuss the models in Appendix \ref{sec:appendixModels}, while
Appendix \ref{sec:appendixPrompts} lists the prompts used in experiments.
Evaluating LLMs~\cite{guo2023evaluating,abdallah2023exploring}, especially for question answering, is challenging due to the verbose nature of the responses. Traditional metrics like Exact Match and F1 score may not be suitable. To address this, we use model-agnostic metrics like Token Recall and Answer String Containment\footnote{\url{https://huggingface.co/spaces/evaluate-metric/squad}}. Token Recall measures how well the model's response covers the ground truth. Answer String Containment assesses if the model's response captures the core answer. 
\begin{table}[tb] 
\begin{adjustbox}{width=0.5\textwidth}

\centering
\begin{tabular}{l|c|c|cccll}
\hline
Method & Parameters & Shots & Precision  & Recall  & F1 & Con  & EM \\ \hline
\multirow{4}{*}{Llama-2} & \multirow{4}{*}{7B}
& 0 & 3.675 & \textbf{33.935} & 6.09 & \textbf{50.315} & 0.035\\
&& 1 & 7.08 & 27.21 & 9.00 & 43.09 & 3.57\\
&& 2 & 23.05 & 30.65 & 23.655 & 33.06 & 21.49\\
&& 3 & \textbf{23.67} & 30.38 & \textbf{24.22} & 31.78 & \textbf{22.25}\\
\hline
\multirow{4}{*}{Llama-2} & \multirow{4}{*}{13B}
& 0 & 3.605 & \textbf{32.76} & 6.05 & \textbf{53.27} & 0.0085\\
&& 1 & 22.865 & 27.91 & 23.345 & 38.11 & 21.525\\
&& 2 & \textbf{31.645} & 32.35 & \textbf{31.56} & 31.465 & \textbf{30.37}\\
&& 3 & 30.37 & 32.03 & 30.495 & 32.46 & 29.00\\
\hline
\multirow{4}{*}{Llama-2} & \multirow{4}{*}{70B}
& 0 & 5.191 & 39.30 & 8.31 & \textbf{57.32} & 0.138\\
&& 1 & 25.74 & 32.61 & 18.8 & 49.60 & 13.61\\
&& 2 & 34.01 & 44.21 & 35.26 & 43.27 & 31.78\\
&& 3 & \textbf{37.09} & \textbf{46.57} & \textbf{38.44} & 42.77 & \textbf{34.59}\\
\hline
\multirow{4}{*}{Mistral-Instruct} & \multirow{4}{*}{7B}
& 0 & 3.73 & \textbf{48.33} & 6.325 & \textbf{58.30} & 0.034\\
&& 1 & 24.68 & 35.33 & 25.87 & 41.81 & 21.83\\
&& 2 & 32.74 & 35.91 & 33.145 & 34.82 & 30.55\\
&& 3 & \textbf{35.28} & 37.91 & \textbf{35.68} & 36.26 & 32.93\\
\hline
\multirow{4}{*}{Mixtral} & \multirow{4}{*}{8x7B}
& 0 & 3.34 & 40.41 & 5.65 & \textbf{60.20} & 0.156\\
&& 1 & 6.76 & 39.06 & 9.17 & 52.87 & 2.75\\
&& 2 & 14.27 & 41.89 & 16.44 & 52.64 & 10.22\\
&& 3 & \textbf{15.49} & \textbf{44.03} & \textbf{17.83} & 51.57 & \textbf{11.25}\\
\hline
\multirow{4}{*}{GPT-3.5} & \multirow{4}{*}{-}
& 0 & 2.68 & 29.12 & 4.56 & \textbf{46.79 }& 0.008 \\
&& 1 &21.12 & 45.58 &  23.95 & 52.77 & 15.80\\
&& 2 & 30.88 & 41.17 & 32.71 & 37.79 & 26.01\\
&& 3 &  \textbf{31.68 }&  \textbf{42.37 }& \textbf{33.40 }& 38.85 & \textbf{26.53}\\
\hline
\multirow{4}{*}{GPT-4o} & \multirow{4}{*}{-}
& 0 & 8.30 & 29.90 & 10.47 & 46.55 & 3.45\\
&& 1 & 21.12 &  45.58 & 23.95 & 52.77 & 15.80\\
&& 2 &  40.82 & 53.83 & 42.92 &  48.12 & 35.62 \\
&& 3 & \textbf{43.91} & \textbf{56.08} & \textbf{45.72} & \textbf{47.62} & \textbf{39.07}\\
\hline
\end{tabular}
\end{adjustbox}
\caption{Performance of few-shot LLM Models.}

\label{tab:fewshot}

\end{table}

\begin{table}[tb] 
\centering
\begin{adjustbox}{width=0.5\textwidth,center}
\begin{tabular}{l|l|l|cccc}
\hline
Method & Parameters & Context & Precision & Recall  & F1 & Con  \\ \hline
\multirow{3}{*}{Llama-2} & \multirow{3}{*}{7B}
& No Context & 3.67 & 33.93 & 6.09 & 50.31 \\
&& Retriever & 3.59 & 33.67 & 5.97 & 53.48 \\
&& True Context & \textbf{3.92} & \textbf{37.40} & \textbf{6.49} & \textbf{56.03} \\
\hline
\multirow{3}{*}{Llama-2} & \multirow{3}{*}{13B}
& No Context & 3.60 & 32.76 & 6.05 & 53.27 \\
&& Retriever & 3.50 & 34.22 & 5.84 & 55.38 \\
&& True Context & \textbf{3.75} & \textbf{37.09} & \textbf{6.28} & \textbf{57.42} \\
\hline
\multirow{3}{*}{Llama-2} & \multirow{3}{*}{70B}
& No Context & 5.19 & \textbf{39.30} & 8.31 & 57.32 \\
&& Retriever & 5.27 & 36.16 & 8.12 & 56.45 \\
&& True Context & \textbf{5.82} & 38.59 & \textbf{8.82} & \textbf{58.26} \\
\hline
\multirow{3}{*}{Mistral-Instruct} & \multirow{3}{*}{7B}
& No Context & 3.73 & \textbf{48.33} & 6.32 & \textbf{58.30} \\
&& Retriever & 3.86 & 33.32 & 6.31 & 55.90 \\
&& True Context & \textbf{5.13} & 35.26 & \textbf{8.08} & 54.14 \\
\hline
\multirow{3}{*}{Mixtral} & \multirow{3}{*}{8x7B}
& No Context & 3.34 & \textbf{40.41} & 5.65 & 60.20 \\
&& Retriever & \textbf{4.23} & 35.93 & \textbf{6.62} & 56.30 \\
&& True Context & 3.65 & 38.02 & 5.88 & \textbf{60.54} \\
\hline

\end{tabular}
\end{adjustbox}
\caption{Performance of LLMs in RAG QA setting.}

\label{tab:RAG}

\end{table}
\subsection{Zero-shot Results}
We conducted zero-shot QA experiments to evaluate different Large Language Models (LLMs). The models generate responses based solely on their pre-training. The results, presented in Table \ref{tab:zero-shotmetrics_llm}, show varying performance across different metrics. Notably, model size is not the sole determinant of performance. For instance, Llama-chat models, with fewer parameters, perform comparably to GPT-3.5. Some models, like GPT-4o, Vicuna and Mistral, suggest a trade-off between precision and comprehensiveness. Models like Zephyr and Falcon, despite lower precision and F1 scores, have high recall and containment scores, indicating their ability to capture significant portions of the ground truth. Lastly, the WizardLM model was found to have lower scores across all metrics.

\begin{table*}[tb] 
\centering
\begin{adjustbox}{width=0.9\textwidth,center}
\begin{tabular}{c|c|cccc|cccc|cccc}
\hline
\multirow{2}{*}{Method} & \multirow{2}{*}{Parameters}  & \multicolumn{4}{c}{Attribute-type}&   \multicolumn{4}{c}{Comparison-type} & \multicolumn{4}{c}{Counting-type}  \\ 

& &  Precision  & Recall & F1  & Con & Precision  & Recall & F1  & Con & Precision  & Recall & F1  & Con   \\ \hline
\multirow{1}{*}{Llama-2} & 7B & 6.55 & 60.70 & 9.52 & 96.86 & 5.30 & 58.05 & 8.83 & 63.20 & 0.34 & 10.50 & 0.66 & 80.24 \\
\multirow{1}{*}{Llama-2} & 13B & 6.17 & 61.26 & 10.67 & 96.77 & 5.22 & 59.53 & 8.75 & 63.30 & 0.16 & 5.86 & 0.30 & 86.13 \\
\multirow{1}{*}{Llama-2} & 70B & 8.21 & 65.85 & 13.71 & \textbf{97.63} & 7.06 & 61.41 & 11.25 & 64.76 & 0.43 & 16.42 & 0.84 & 48.99 \\
\multirow{1}{*}{Mistral-Instruct} & 7B & 5.87 & 54.00 & 9.96 & 96.52 & 8.33 & 57.79 & 12.73 & 62.18 & 0.68 & 20.80 & 1.31 & 87.45 \\
\multirow{1}{*}{Mixtral} & 8*7B& 4.04 & \textbf{75.80} & 7.21 & 97.25 & 5.18 & 63.50 & 8.86 & \textbf{72.10} & 0.60 & \textbf{27.44} & 1.17 & \textbf{90.10} \\
\multirow{1}{*}{GPT-3.5} & - & 9.48 &73.16 &15.87 & 70.02 &6.51  & 55.07& 10.41 & 43.50  & 0.20 & 4.84  &  0.39 & 77.38\\

\multirow{1}{*}{GPT-4o} & -& \textbf{23.94} &75.56 &\textbf{30.05} &72.66 & \textbf{12.38} &\textbf{71.63} &  \textbf{17.77}& 57.17  &\textbf{2.16} &  20.67 &  \textbf{3.46} & 78.61\\
\hline

\end{tabular}
\end{adjustbox}
\caption{Performance on Attribute-, Comparison-, and Counting-type questions.}

\label{tab:example1}
\end{table*}

\begin{table*}[tb] 
\centering
\vspace{-0.2cm}
\begin{adjustbox}{width=0.9\textwidth,center}
\begin{tabular}{c|c|cccc|cccc|cccc}
\hline
\multirow{2}{*}{Method} & \multirow{2}{*}{Parameters}  & \multicolumn{4}{c}{Entity}&   \multicolumn{4}{c}{Event} & \multicolumn{4}{c}{Time}  \\ 

& &  Precision  & Recall & F1  & Con & Precision  & Recall & F1  & Con & Precision  & Recall & F1  & Con   \\ \hline
\multirow{1}{*}{Llama-2} & \multirow{1}{*}{7B} & 4.26 & 40.86 & 6.95 & 75.80 & 4.75 & 53.25 & 7.87 & 82.00 & 3.38 & 41.86 & 5.84 & 76.09 \\
\multirow{1}{*}{Llama-2} & \multirow{1}{*}{13B} & 3.79 & 40.70 & 6.52 & 80.95 & 4.96 & 49.72 & 8.31 & 81.80 & 3.14 & 38.27 & 5.45 & 81.28 \\
\multirow{1}{*}{Llama-2} & \multirow{1}{*}{70B} & 5.07 & 47.55 & 8.45 & 82.31 & 7.23 & 60.18 & 11.28 & 85.10 & 4.50 & 53.50 & 7.67 & 84.02 \\

\multirow{1}{*}{Mistral-Instruct} & \multirow{1}{*}{7B} & 3.91 & 35.43 & 6.53 & 79.20 & 7.49 & \textbf{62.28} & 11.64 & 84.60 & 4.31 & 39.85 & 6.95 & 81.55 \\
\multirow{1}{*}{Mixtral} & \multirow{1}{*}{8*7B} & 2.82 & 50.95 & 4.85 & \textbf{84.76} & 4.91 & 60.06 & 8.29 & \textbf{86.60} & 2.62 & 54.78 & 4.77 & \textbf{85.85} \\
\multirow{1}{*}{GPT-3.5} & -& 5.59 & 43.64 &9.32 & 64.47 & 6.80 & 52.17&  10.73 & 58.63  & 4.33 & 39.23  & 7.32  & 64.63\\

\multirow{1}{*}{GPT-4o} & - & \textbf{14.10} &\textbf{ 51.55} & \textbf{17.10} & 72.03 &\textbf{ 13.77} & 57.17 & \textbf{18.02} & 59.76 &  \textbf{11.00} & \textbf{57.12} & \textbf{15.85} &  72.78\\
\hline

\end{tabular}
\end{adjustbox}
\caption{Performance on Entity, Event, and Time questions.}

\label{tab:example2}
\end{table*}

\subsection{Few-shot Results}
In the few-shot learning setting ~\cite{chada2021fewshotqa}, models improve as they are provided with more examples, as seen in Table \ref{tab:fewshot}. Across all models, performance increases with additional shots, but the rate of improvement plateaus after two shots, indicating diminishing returns. The Llama-2 7B, 13B, and 70B models exhibit steady gains, with the 70B variant achieving the highest performance among them. Similarly, Mistral-Instruct and Mixtral models follow the same trend, though with smaller absolute gains. Notably, GPT-4o outperforms all models, showing a significant improvement in F1 score from 10.47 (zero-shot) to 45.72 (three-shot), along with the highest recall and containment scores, demonstrating its superior ability to adapt with few-shot examples. Finally, as expected, GPT-3.5's performance is much worse than the one of GPT-4o.

\subsection{RAG Results}

Retrieval-augmented generation (RAG)~\cite{lewis2020retrieval,abdallah2023generator} combines the strengths of pre-trained language models and information retrieval systems to generate responses in a question-answering setting. In RAG, when a question is posed, relevant documents are first retrieved from a large corpus. These retrieved documents are then provided as additional context to a language model, which generates a response based on both the original question and the retrieved documents. Following ~\cite{karpukhin2020dense}, we use the English Wikipedia dump from Dec. 20, 2018 as the source documents for answering questions, which contains 21,015,324 passages in total. Each passage is prepended with the title of the Wikipedia article from which it originates, along with an [SEP] token. 

In this experiment, we aimed to evaluate the efficiency of appending the retrieved context to LLMs retrieved by Dense Passage Retriever (DPR)~\cite{karpukhin2020dense} for question answering. We tested three different settings for each model: \emph{without context}, \emph{with the first top retrieved passage} as context, and \emph{with the true context}. The true context is determined by retrieving $1,000$ passages using DPR and conducting a simple search within these passages. If the answer was found within a passage, we selected the first passage that contains the answer as the true context. If the answer was not found in any of the retrieved passages, we selected a random passage.

Table \ref{tab:RAG} presents the results. The performance of the models usually improves when context is provided, with the true context generally leading to the best performance. This suggests that providing relevant context can help guide the models in generating more accurate and relevant responses. However, the performance varies across different models and settings, indicating that the effectiveness of RAG depends on both the specific model and the quality of the retrieved context.

\subsection{Results on Different Question Types}
Across different question types—attribute, comparison, counting, entity, event, and time—the performance of LLMs varies significantly, as shown in Tables \ref{tab:example1} and \ref{tab:example2}. Counting type questions are most challenging followed by the comparison questions and then the attribute type questions. 

Llama 
models exhibit a steady improvement with increasing parameters, with Llama-2 70B outperforming its smaller variants across most metrics. However, Mistral-Instruct and Mixtral, despite having fewer parameters, achieve comparable or better results in certain cases, particularly in recall and containment scores. GPT-4o consistently delivers the highest performance across all categories, achieving the best precision, recall, and F1 scores, particularly excelling in attribute and comparison-type questions. It also dominates in entity, event, and time-based questions, highlighting its strong generalization ability. GPT-3.5 performs well but falls behind GPT-4o, particularly in recall and containment, indicating a weaker ability to retrieve and structure temporal knowledge.

\section{Conclusions}

We introduced \textsc{ComplexTempQA}, a large-scale dataset comprising over 100 million temporal question-answer pairs, surpassing existing benchmarks in scope, coverage, and complexity. Built on Wikipedia and Wikidata, it spans more than 30 years and covers a wide range of domains, including history, politics, sports, and science.
We introduced a taxonomy categorizing questions into three key types: \textit{attributes}, \textit{comparisons}, and \textit{count}, each requiring advanced temporal knowledge and temporal reasoning skills such as multi-hop inference, temporal aggregation, and event ordering. To support targeted evaluation, each question is enriched with structured metadata, enabling precise assessment of LLMs’ ability to process and reason over temporal information. In Appendix \ref{sec:appendixdatasetUse} we discuss the different use cases of our dataset.

Finally, we evaluated several LLMs, revealing significant gaps in their capabilities. While state-of-the-art models performed relatively well on simpler questions, their performance on more complex temporal questions decreased significantly, 
highlighting the challenging character of our dataset.
\section{Limitations: } 
Despite its advantages, \textsc{ComplexTempQA} has several limitations. The dataset is built upon Wikipedia and Wikidata, which are characterized by relatively high precision but may suffer from lower recall, meaning that while available facts are generally accurate, relevant historical or domain-specific facts might be missing. The dataset is also constrained by its timeframe, as it primarily covers the period of 1987 until 2023,  limiting its applicability to broader historical analysis. Additionally, the temporal scope of questions may not always align perfectly with evolving real-world knowledge, as both Wikipedia and Wikidata are continuously updated. This is also a challenge in our RAG analysis for which we employ the Wikipedia dump from 2018 which is however commonly used as a retrieval corpus in RAG studies; thus we have adapted the same setting in our experiments. 
Moreover, the dataset contains a significant proportion of comparative questions, which, while valuable for evaluating comparative reasoning over historical knowledge, may introduce a bias towards one form of temporal inference. Addressing these challenges could improve the dataset’s utility in future iterations.

\appendix

\bibliography{acl_2025}

\section{Example Questions and Templates}
\label{sec:appendixTemplates}

In our question generation process, we employed a range of templates, each tailored to different contexts and requirements. Below, we present several selected question templates showcasing the diversity of data:

\noindent
\textbf{Attribute Queries}
\begin{itemize}
    \item \texttt{What was [ATTR] of [ENTITY]?}\\
    \textit{Example: What was the genre of the movie \emph{Border}?}

    \item \texttt{When was [PERSON] [ENTITY]?}\\
    \textit{Example: When was Girija Prasad Koirala President of Nepal?}

        \item \texttt{When [V1] [V2] [ENTITY]?}\\
    \textit{Example: When was the publication date of the movie in which Lou Diamond Phillips and Esai Morales acted?}
\end{itemize}

\noindent
\textbf{Comparison Queries}
\begin{itemize}
    \item \texttt{Comparing [P1] of [ENTITY1] and [ENTITY2], which one has a [COMPARE] [P2]?}\\
    \textit{Example: Comparing the country of the company Google and the company Yandex, which one has a lower highest point?}

    \item \texttt{Did [EVENT1] have a [COMPARE] [ATTR] than [EVENT2]?}\\
    \textit{Example: Did the car bombing in 1993 (6 deaths) have a higher death toll than the train wreck in 1989 (645 deaths)?}
    
    \item \texttt{Did [EVENT1] or [EVENT2] happen first?}\\
    \textit{Example: Did Helios Airways Flight 522 accident or the 27th G8 summit happen first?}

    \item \texttt{Which one happened first, [EVENT1], [EVENT2], or [EVENT3]?}\\
    \textit{Example: Which happened first, the 69th Academy Awards, the USS Cole bombing, or the Daegu subway fire?}

    \item \texttt{Did [EVENT1] happen [SIGNAL] [EVENT2]?}\\
    \textit{Example: Did the Gulf War happen after Hurricane Hugo?}

\end{itemize}

\noindent
\textbf{Counting Queries}
\begin{itemize}
    \item \texttt{How many times did [PERSON] win [ENTITY] [TIME]?}\\
    \textit{Example: How many times did Robert Richardson win the Academy Award for Best Cinematography between 1987 and 2007?}

    \item \texttt{How often did [EVENT] happen [TIME]?}\\
    \textit{Example: How often did an aviation accident with more than 54 participants in Argentina happen in 1999?}

    \item \texttt{In how many years did [EVENT] [YEAR] happen?}\\
    \textit{Example: In how many years did a aviation accident with more than 28 survivors between 2008 and 2023 by Xiamen Airlines happen?}
\end{itemize}

\noindent
\textbf{Multi-hop Queries}
\begin{itemize}
    \item \texttt{Did [V] [ENTITY1] [SIGNAL] [V] [ENTITY2]?}\\
    \textit{Example: Did the publication of the movie with Lou Diamond Phillips and Esai Morales happen before the publication of \emph{Resident Evil: Extinction}?}

    \item \texttt{What was the [P2] of the [P1] of [ENTITY] in meters?}\\
    \textit{Example: What was the highest point (in meters) of the country where the Moscow theater hostage crisis happened?}
\end{itemize}

We employ placeholders to convey specific elements within the question templates. The placeholder \texttt{[SIGNAL]} denotes temporal relationships, such as ''before'' or ''after,'' indicating the sequence of events or entities. \texttt{[ATTR]} stands for attributes like the number of injured people, providing context or additional information related to the events or entities. \texttt{[YEAR]} referring either to a specific year or between two years e.g. \textit{''in 1987''} or \textit{''between 1987 and 2007''}. \texttt{[COMPARE]} signifies comparison relationships, such as ''higher'' or ''lower,'' enabling the comparison of attributes or characteristics between events, entities, or their attributes. Then we have \texttt{[V]} to denote various verbs like ''was'' or ''happened''. \texttt{[P]} stands for properties used to create multi hop questions. Furthermore, we make a clear distinction between questions involving persons and those involving other entities, as the question structure may vary accordingly. Despite these differences, some question templates are versatile enough to accommodate both event and entity questions, ensuring flexibility and adaptability in our approach to question generation.

\section{Metadata}

We provide the following metadata:\footnote{For the Wikidata IDs we exclude the leading \textit{'Q'} as well for the properties the leading \textit{'P'}}

\textbf{Entity in question:} A list of Wikidata IDs of the question.

\textbf{Entity in answer:} A list of Wikidata IDs of the answer if it contains an entity.

\textbf{Country in question:} A list of country Wikidata IDs of the countries of the questioned entities.

\textbf{Country in answer:}  A list of country Wikidata IDs of the countries of the entities in the answer.

\textbf{Hop property:} A list of Wikidata properties of the question if it contains a hop. If there are multiple hops, they are listed in the order of use.

\textbf{Rating:} A numerical rating where \textit{0} is considered a popular (or easy) question, and \textit{1} is considered a less popular or harder question according to the rating as described in Section \ref{sec:Rating}.

\textbf{Is unnamed:} A numerical which is \textit{1} if the question contains an implicitly described event and \textit{0} otherwise.

\textbf{Type:} The type based on the taxonomy given in Figure \ref{fig:taxonomy}.

\textbf{Time span:} The time frame to which the question relates to. For example, for the entity questions, the time frame ranges from born/creation to death/destruction. The start is always the earlier date, and the end is the latter.

Below is an example for a question including the metadata:

\begin{itemize}
    \item \textbf{Question}: What was the highest point of the country where the 1988 Summer Olympics happened, in meters?
    \item \textbf{Answer}: [1950]
    \item \textbf{Entity in question}: [8470]
    \item \textbf{Entity in answer}: []
    \item \textbf{Country in question}: [884]
    \item \textbf{Country in answer}: []
    \item \textbf{Hop property}: [17, 610]
    \item \textbf{Rating}: 1
    \item \textbf{Is unnamed}: 0
    \item \textbf{Time span}: [1988-09-17, 1988-10-02]
\end{itemize}
\vspace{-0.2cm}

\section{Models used in Experiments} 
\label{sec:appendixModels}
In experiments we test multiple Large Language Models, including Llama, Mistral, Mixtral, Falcon, Vicuna, Zephyr, WizardLM, GPT-3.5, and GPT-4-o, each selected for its unique strengths.
Llama models, developed by Meta~\cite{touvron2023llama}, leverage reinforcement learning with human feedback (RLHF) for dialogue optimization. Mistral-7B and Mixtral (Sparse Mixture of Experts) outperform Llama-2 13B and 70B, respectively, in various benchmarks~\cite{jiang2023mistral}. GPT-3.5 improves upon GPT-3 by reducing toxicity and enhancing contextual understanding~\cite{brown2020language}. Falcon, optimized for inference, utilizes multi-query attention and FlashAttention~\cite{dao2022flashattention}. Vicuna fine-tunes Llama-2 using ShareGPT data, enhancing conversational capabilities, while Zephyr employs distilled supervised fine-tuning (dSFT) for improved task accuracy~\cite{tunstall2023zephyr}. Finally, WizardLM, based on Llama-2 13B, refines instruction-following abilities using an Evol-Instruct method~\cite{xu2023wizardlm}.

\section{Experiment Prompts} 
\label{sec:appendixPrompts}

In all experiments, each LLM was prompted to function as a helpful assistant and deliver direct, concise answers. For models employing the Retrieval Augmented Generation (RAG) approach, the prompt included additional context from retrieved documents to enhance the response quality. This modification ensured that responses were informed by relevant background information.

 For all the experiments involving LLMs, we used the following prompt:
 \begin{quote}
 You are a helpful assistant. Provide direct and concise answer to the following question. \\
 Question: {<question>.}
 \end{quote}
 In the case of RAG, the prompt is slightly modified to incorporate the additional context provided by the retrieved documents. The prompt used for RAG is:
 \begin{quote}
 You are a helpful assistant. Using the context, provide direct and concise answers to the following question. \\
 Question: {<question>}.\\
 Context: {<context>}.
 \end{quote}

\section{Dataset Use}
\label{sec:appendixdatasetUse}
We briefly list below the intended use cases of our dataset.

\textbf{LLM Evaluation and Training.} \textsc{ComplexTempQA} is an effective resource for evaluating LLMs, as demonstrated in our preliminary study (Sec. 6). As the largest QA dataset currently available, it provides an unparalleled foundation for analyzing LLM performance. It supports fine-tuning, prompt engineering, and the assessment of temporal question answering capabilities~\cite{wallat}. Notably, the dataset facilitates rigorous evaluation of truthfulness by offering unprecedented diversity and scale—critical factors for mitigating hallucinations. Moreover, its rich temporal metadata expands the scope of time-based QA research~\cite{Event-QA}.

\textbf{Continual Learning and Adaptation of LLMs.} The detailed temporal annotations and extensive scale of \textsc{ComplexTempQA} make it useful for online adaptation and continual training approaches \cite{hu2023meta,tack2024online}. With approximately 280k questions per year on average, it enables targeted experiments on temporal adaptation—vastly outperforming benchmarks like ArchivalQA~\cite{ArchivalQA}, which offers only around 22k questions per year.

\textbf{RAG Systems.} \textsc{ComplexTempQA} can be used to train and evaluate Open Domain Question Answering models with historical news archives, such as the NYT Annotated Archive~\cite{sandhaus2008new} (1.8 million articles from 1987 to 2007). Given the NYT's international coverage and our focus on US events, most questions can be answered using this archive, making our dataset a complementary resource for temporal IR research~\cite{ArchivalQA,campos2014survey,DBLP:conf/sigir/WangJYC23}.

\textbf{KGQA Systems.} Built from Wikidata and Wikipedia, \textsc{ComplexTempQA} is well-suited for Knowledge Graph Question Answering~\cite{usbeck20177th,souza2020event}. Its integration with large-scale knowledge graphs such as Wikidata and SemOpenAlex~\cite{SemOpenAlex}—which contain billions of facts—enhances QA models’ ability to explore complex temporal relationships and evaluate multi-hop reasoning~\cite{han2023generating}.

\end{document}